\documentclass[twoside]{article}

\usepackage[accepted]{aistats2024}

\usepackage[round]{natbib}

\usepackage{amsfonts}
\usepackage{amsmath}
\usepackage{amsthm}
\usepackage{algorithm}
\usepackage{algorithmic}
\usepackage{booktabs}
\usepackage[pdftex]{graphicx}
\usepackage{pdftexcmds}
\usepackage{xcolor}
\usepackage{subcaption}

\DeclareFontFamily{U}{cbgreek}{}
\DeclareFontShape{U}{cbgreek}{m}{n}{
        <-6>    grmn0500
        <6-7>   grmn0600
        <7-8>   grmn0700
        <8-9>   grmn0800
        <9-10>  grmn0900
        <10-12> grmn1000
        <12-17> grmn1200
        <17->   grmn1728
      }{}
\DeclareFontShape{U}{cbgreek}{bx}{n}{
        <-6>    grxn0500
        <6-7>   grxn0600
        <7-8>   grxn0700
        <8-9>   grxn0800
        <9-10>  grxn0900
        <10-12> grxn1000
        <12-17> grxn1200
        <17->   grxn1728
      }{}

\DeclareRobustCommand{\Qoppa}{%
  \text{\usefont{U}{cbgreek}{\normalorbold}{n}\symbol{21}}%
}
\makeatletter
\newcommand{\normalorbold}{%
  \ifnum\pdf@strcmp{\math@version}{bold}=\z@ bx\else m\fi
}
\makeatother

\def\Var{{\mathrm{Var}}}
\def\Cov{{\mathrm{Cov}}}

\makeatletter
\newtheorem*{rep@theorem}{\rep@title}
\newcommand{\newreptheorem}[2]{%
\newenvironment{rep#1}[1]{%
 \def\rep@title{#2 \ref{##1}}%
 \begin{rep@theorem}}%
 {\end{rep@theorem}}}
\makeatother

\newtheorem{theorem}{Theorem}
\newreptheorem{theorem}{Theorem}
\newtheorem{corollary}{Corollary}[theorem]
\newtheorem{definition}{Definition}

\begin{document}

\twocolumn[

\aistatstitle{An Analytic Solution to Covariance Propagation in Neural Networks}

\aistatsauthor{ Oren Wright \And Yorie Nakahira \And José M. F. Moura }

\aistatsaddress{ Carnegie Mellon University \And Carnegie Mellon University \And Carnegie Mellon University } ]

\begin{abstract}
Uncertainty quantification of neural networks is critical to measuring the reliability and robustness of deep learning systems. However, this often involves costly or inaccurate sampling methods and approximations. This paper presents a sample-free moment propagation technique that propagates mean vectors and covariance matrices across a network to accurately characterize the input-output distributions of neural networks. A key enabler of our technique is an analytic solution for the covariance of random variables passed through nonlinear activation functions, such as Heaviside, ReLU, and GELU. The wide applicability and merits of the proposed technique are shown in experiments analyzing the input-output distributions of trained neural networks and training Bayesian neural networks.

\end{abstract}

\section{INTRODUCTION}

 This paper presents an analytic moment propagation technique to accurately characterize the input-output distributions of deep neural networks. Neural networks have achieved state-of-the-art inference accuracy across many problem domains, but their application to safety-critical domains like autonomous vehicles, industrial robots, and medicine demands not only high accuracy but also high reliability and robustness. Accurate uncertainty quantification in neural networks enables measurements of reliability and robustness, since uncertainty quantification gives us a language in which we can reason about the likelihood and conditions of failure. Uncertainty quantification techniques can be used to analyze the robustness of trained neural networks to input noise and adversarial attacks, or to train networks that represent predictive uncertainty to provide well-calibrated confidence scores and out-of-distribution detection. 
 
 Rather than explicitly representing uncertainty, neural networks are commonly trained to produce point estimates without rigorous uncertainty quantification. Although their optimization process has a probabilistic interpretation---e.g., minimizing a cross-entropy loss function corresponds to maximum likelihood estimation under certain conditions \citep{mackay2002book}---networks trained in this way are often poorly calibrated \citep{guo_calibration_2017} and can perform poorly against small input perturbations and adversarial attacks \citep{szegedy2014, carlini2017}.

 One approach to quantifying uncertainty in neural networks is by propagating statistical moments across a network, from input to output, layer by layer. In the existing literature, moment propagation has been used 1) to analyze the input-output distributions of any given (or well-trained) networks~\citep{bibi2018}; 2) to train probabilistic networks that explicitly model input uncertainty~\citep{gast2018lightweight}; and 3) to train fully Bayesian neural networks (BNNs) that model both input uncertainty and parameter uncertainty~\citep{wu2019}. 

 Uncertainty quantification remains a challenging area of research, because modern neural networks are nonlinear and represent data in high-dimensional spaces. For moment propagation in particular, interactions between the outputs of activation functions are challenging to characterize analytically. The moment propagation techniques proposed in prior work approximate the covariance between activation variables, as no analytic solution for covariance has been published for generalized neural network nonlinearities.
 
 In this paper, we address the above problem by deriving an exact analytic solution for the covariance matrix of nonlinear activation functions with Gaussian inputs. Our covariance theorem, presented in Section~\ref{sec:moment_prop}, is a general solution that can be computed to arbitrary precision, enabling more accurate and widely applicable moment propagation in neural networks. As illustrative examples, we show how our solution computes the covariance for neural networks with Heaviside, rectified linear unit (ReLU), and Gaussian error linear unit (GELU) activation functions. With these examples we also derive other activation statistics missing from the current literature. In Section~\ref{sec:experiments}, we demonstrate the application of these theoretical results to both the analysis and synthesis of neural networks, showing in experiments improved accuracy for characterizing input-output distributions of trained neural networks and for training Bayesian neural networks (see Table~\ref{tab:uci}). Code to run these experiments is publicly available\footnote{https://github.com/omwright/cov-prop-nn}.
 
\begin{figure}[!t]
\includegraphics[width=3.2in]{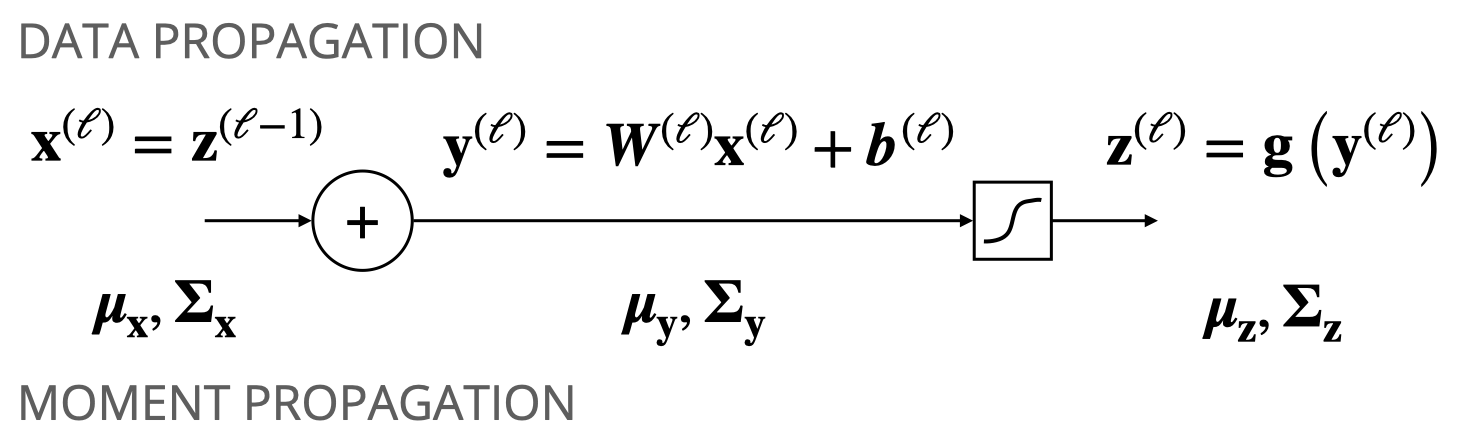}
\caption{Moment propagation for a single network layer with an affine transformation followed by a nonlinear activation function. Layer-by-layer mean and covariance statistics can be used as a measure of uncertainty traveling across a network.}
\label{fig:network_layer}
\end{figure}

\section{RELATED WORK}\label{sec:related_work}
Below we discuss related work in the quantification of neural network uncertainty.
\subsection{Uncertainty Propagation}
The earliest work on propagating uncertainty through neural networks appears in research on Bayesian belief networks~\citep{pearl1988, neal1992belief}. \cite{frey_variational_1999} propose a framework for transformations of Gaussian variables in nonlinear Gaussian belief networks. \cite{gast2018lightweight} build upon this framework and propose a method to propagate mean and variance statistics for training neural networks with assumed density filtering~\citep{boyen1998adf, minka2001} and probabilistic output layers. This prior work models the predictive uncertainty of a neural network, but is limited by the fact that only variance, not covariance, is propagated (i.e., only the diagonal of a covariance matrix), due to the lack of exact covariance solutions.

Another application of uncertainty propagation is the measurement of adversarial robustness, which is commonly done by evaluating adversarial accuracy, i.e., the network test accuracy under selected adversarial attacks. Evaluating adversarial accuracy is usually done with attacks that have proven empirically effective, such as those proposed in \cite{carlini2017} and \cite{madry2018pgd}. However, because one must select and implement a particular set of attacks, adversarial accuracy is a coarse and sometimes misleading measure of robustness \citep{ding2019sensitivity, olivier23perturbations}. \cite{bibi2018} propose a technique to propagate moments through ReLU networks via piecewise-linear approximations (PL-DNN) to give a probabilistic model of noise and adversarial robustness, which they use to measure adversarial fooling rates and pixel-level noise sensitivity in computer vision problems. The authors give a closed-form solution for the covariance of Gaussian random variables passed through a ReLU activation, but only for zero-mean Gaussians.

In other work, \cite{astudillo11_interspeech} and \cite{abdelaziz2015interspeech} combine approximations with sampling-based methods to propagate uncertainty through networks for automatic speech recognition. \cite{daunizeau_semi-analytical_2017} derives accurate approximations for the mean and variance of the sigmoid and softmax functions. \cite{seo_distribution_2021} show how the distribution of the penultimate layer of a neural network can be used for generative modeling and knowledge distillation.

\subsection{Bayesian Neural Networks}\label{sec:related_bnn}
Bayesian inference with neural networks was first proposed by \cite{mackay1992}, modeling both input and parameter uncertainty. Because a BNN computes a full posterior distribution, it is robust to overfitting and can quantify predictive uncertainty. However, because modern neural networks are high-dimensional and non-convex, computing posterior distributions incurs a prohibitive computational cost and typically requires approximative, sampling-based methods \citep{graves2011variational, blundell2015weightuncertainty, hernandez-lobato2015pbp, kingma2015reparam}.

Due to this cost and difficulty, some approximative methods avoid directly representing the uncertainty of the parameters \citep{gal2016dropout, maddox2019swag}. \cite{loquercio2020uncertainty} combine the assumed density filtering approach of \cite{gast2018lightweight} with the dropout sampling of \cite{gal2016dropout} and apply their framework to autonomous systems. Most closely related to our work is \cite{wu2019}, which proposes an alternative, sample-free approach to Bayesian inference, deterministic variational inference (DVI), which analytically propagates the mean and covariance statistics of inputs, activations, and parameters. To compute the activation covariance matrix, the authors propose closed-form approximations of the covariance of Heaviside and ReLU activations. \cite{MAE2021394} build on the approximations of \cite{daunizeau_semi-analytical_2017} and \cite{wu2019} to compute uncertainty for dropout-based Bayesian neural networks. The reliance on activation-specific covariance approximations or variance alone limits the accuracy and applicability of these approaches.

\section{MOMENT PROPAGATION}\label{sec:moment_prop}

In this section we describe moment propagation in neural networks and propose an analytic solution. We denote scalars, vectors, and matrices by $x$, $\mathbf{x}$, and $\mathbf{X}$, respectively. We do not distinguish between random variables and their instantiations, in part because we may treat a given variable as random or deterministic depending upon the context. For economy of notation, we avoid subscripts and superscripts except when necessary.

\subsection{Problem Formulation}\label{sec:prob_formulation}
Given some uncertainty at the input to a neural network, whether due to noise, adversarial perturbations, or otherwise---or possibly at some intermediate stage of the network---our goal is to compute the statistics or probability distribution of the network output. More explicitly, we treat inputs probabilistically and propagate their mean and covariance layer by layer, as depicted in Figure~\ref{fig:network_layer}, and use these moments to determine output (or intermediate) uncertainty. Each layer $\ell$ in a standard multilayer perceptron takes an input $\mathbf{x}^{(\ell)} \in \mathbb{R}^n$ and is composed of an affine transformation 
\begin{equation*}
    \mathbf{y}^{(\ell)} = \mathbf{W}^{(\ell)}\mathbf{x}^{(\ell)} + \mathbf{b}^{(\ell)}
\end{equation*}
followed by a nonlinear, element-wise activation function
\begin{equation*}
    \mathbf{z}^{(\ell)} = \mathbf{g}^{(\ell)}\left(\mathbf{y}^{(\ell)}\right).
\end{equation*}
The output of the activation function then serves as the input to the affine transformation of the next layer, $\mathbf{x}^{(\ell + 1)} = \mathbf{z}^{(\ell)}$.

As we will show, we can compute layer-wise mean and covariance across this structure under certain assumptions, giving us a general, sample-free framework for propagating uncertainty through neural networks. This can be used to analyze the robustness of trained networks \citep{bibi2018}, or to train probabilistic neural networks that can represent predictive uncertainty \citep{hernandez-lobato2015pbp, gast2018lightweight, wu2019}.

\subsection{Activation Covariance}

To make the problem tractable, we assume that the input to each nonlinearity is Gaussian, $\mathbf{y} \sim \mathcal{N}(\boldsymbol{\mu}_\mathbf{y}, \mathbf{\Sigma}_\mathbf{y})$. \cite{hernandez-lobato2015pbp, bibi2018, gast2018lightweight, wu2019} make similar Gaussian assumptions in their moment propagation algorithms. \cite{hernandez-lobato2015pbp, gast2018lightweight} use assumed density filtering~\citep{boyen1998adf, minka2001}, which requires parametric distributions like the Gaussian. \cite{wu2019} observe that, empirically, for the high dimensions and conditions typical of practical neural networks, the Gaussian assumption can be loosely justified by the central limit theorem, regardless of the shape of the distribution of $\mathbf{x}$. The Gaussian assumption may also be justified from the principle of maximum entropy~\citep{jaynes03}.

Even if the distribution of the activation input $\mathbf{y}$ is Gaussian, computing the moments of the activation output $\mathbf{z}$ is non-trivial. For the covariance matrix $\mathbf{\Sigma_z}$ in particular, no exact solution exists in the literature, and existing methods use activation-specific approximations (see, e.g., \cite{bibi2018, wu2019}). We present a general method to compute the covariance between functions of Gaussian random variables below in Theorem \ref{thm:cov}, a complete proof of which can be found in Appendix~\ref{sec:appendix_cov}.

\begin{definition}\label{thm:def}
A vector function $\mathbf{g}: \mathbb{R}^{n} \rightarrow \mathbb{R}^{n}$ is \emph{element-wise independent and identical} if
\begin{equation*}
    \mathbf{g}(\mathbf{y}) = \begin{bmatrix}
        g(y_1) & g(y_2) & \ldots & g(y_n)
    \end{bmatrix}^\top
\end{equation*}
with $g: \mathbb{R} \rightarrow \mathbb{R}$.
\end{definition}
\begin{theorem}
\label{thm:cov}
Suppose $\mathbf{y} \in \mathbb{R}^{n}$ is a multivariate Gaussian random vector, $\mathbf{y} \sim \mathcal{N}(\boldsymbol{\mu}, \mathbf{\Sigma)}$, and $\mathbf{z} = \mathbf{g}(\mathbf{y})$ is an element-wise independent and identical function of $\mathbf{y}$. If the mean and variance of any input element $y_i$ is $(\mu_i, \sigma^2_i)$, and any two input elements $y_i, y_j$ are related by Pearson correlation coefficient $\rho_{ij} = \mathbb{E}[(y_i - \mu_i)(y_j - \mu_j)]/(\sigma_i \sigma_j)$, then the covariance between output elements $z_i$ and $z_j$ is
\begin{equation*}
    \Cov(z_i, z_j) = \sum^{\infty }_{k=1} \frac{\rho^{k}_{ij}}{k!}
    \biggl(
  	\sigma_{i}^{k}\ \frac{\partial^{k} \mathbb{E}[z_i] }{\partial \mu^{k}_{i}}
    \biggr)
    \biggl(
  	\sigma_{j}^{k}\ \frac{\partial^{k} \mathbb{E}[z_j] }{\partial \mu^{k}_{j}}
    \biggr).
\end{equation*}
\end{theorem}

\begin{corollary}\label{thm:var}
Given $z = g(y)$, a function of a Gaussian random variable $y \sim \mathcal{N}(\mu, \sigma^2)$, the variance of $z$ is
\begin{equation*}
    \Var(z) = \sum^{\infty }_{k=1} \frac{1}{k!}
    \left(
 	\sigma^{k}\ \frac{\partial^{k} \mathbb{E}[z] }{\partial \mu^{k}} 
    \right)^2.
\end{equation*}
\end{corollary}

Theorem~\ref{thm:cov} is a general solution that can be computed to arbitrary precision. Computing the covariance matrix of $\mathbf{z} \in \mathbb{R}^n$ with a $k$-th order expansion of Theorem~\ref{thm:cov} has an asymptotic runtime complexity of $O(kn^2)$. Theorem~\ref{thm:cov} is not limited to any particular activation function, assuming only the Gaussianity of the input and that the activation mean and its derivatives are well-defined. As we show below in Section~\ref{sec:prop_activations}, we can readily apply Theorem~\ref{thm:cov} to several common activation functions and successfully propagate across network layers.

\subsection{Propagation Across a Network Layer}\label{sec:prop_activations}

The input and output of a neural network layer are related by
\begin{align}
    \mathbf{y} &= \mathbf{W}\mathbf{x} + \mathbf{b} \\
    \mathbf{z} &= \mathbf{g}(\mathbf{y})
\end{align}
where $\mathbf{g}$ takes the element-wise independent and identical form given in Definition~\ref{thm:def}. If the input $\mathbf{x}$ has mean $\boldsymbol{\mu_x}$ and covariance $\mathbf{\Sigma_x}$, the mean and covariance of $\mathbf{y}$ are
\begin{align}
    \boldsymbol{\mu}_{\mathbf{y}} &= \mathbf{W}\boldsymbol{\mu}_{\mathbf{x}} + \mathbf{b} \label{eq:mean_y} \\
    \mathbf{\Sigma_y} &= \mathbf{W}\mathbf{\Sigma_x} \mathbf{W}^\top. \label{eq:cov_y}
\end{align}

We next compute the mean and covariance of $\mathbf{z}$ in terms of the mean and covariance $\mathbf{y}$, which, together with Equations~\eqref{eq:mean_y} and~\eqref{eq:cov_y} relate the output moments to the input moments.

For simplicity, consider from $\mathbf{z}$ a single element, $z = g(y)$, $y \sim \mathcal{N}(\mu, \sigma^2)$. With the standard normal Gaussian probability density function and standard normal cumulative density function
\begin{align*}
\phi(x) &= \frac{1}{\sqrt{2 \pi}} \exp \left\{- \frac{x^2}{2} \right\} \\
\Phi(x) &= \frac{1}{2} \left[ 1 + \mathrm{erf}\left( \frac{x}{\sqrt{2}} \right) \right]
\end{align*}
the mean of the output for Heaviside, ReLU, and GELU activation functions can be expressed compactly as
\begin{align}
    \text{Heaviside} \qquad \mathbb{E}[z] &= \Phi \left( \frac{\mu}{\sigma} \right) \label{eq:mean_heaviside} \\
    \text{ReLU} \qquad \mathbb{E}[z] &= \mu\Phi \left( \frac{\mu}{\sigma } \right)  +\sigma \phi \left( \frac{\mu}{\sigma } \right) \label{eq:mean_relu} \\
    \text{GELU} \qquad \mathbb{E}[z] &= \mu \Phi \left( \frac{\mu}{\sqrt{1 + \sigma^2}} \right) + \nonumber \\ 
    &\frac{\sigma^2}{\sqrt{1 + \sigma^2}} \phi \left( \frac{\mu}{\sqrt{1 + \sigma^2}} \right) \label{eq:mean_gelu}
\end{align}
and the first derivative with respect to $\mu$ is
\begin{align}
    \text{Heaviside} \qquad \frac{\partial \mathbb{E}[z] }{\partial \mu} &= \frac{1}{\sigma} \phi \left( \frac{\mu}{\sigma} \right) \label{eq:deriv_heaviside} \\
    \text{ReLU} \qquad \frac{\partial \mathbb{E}[z] }{\partial \mu} &= \Phi \left( \frac{\mu}{\sigma} \right)  \label{eq:deriv_relu} \\
    \text{GELU} \qquad \frac{\partial \mathbb{E}[z] }{\partial \mu} &= \Phi \left( \frac{\mu}{\sqrt{1 + \sigma^2}} \right) + \nonumber \\ 
    &\frac{1}{(1 + \sigma^2)^{3/2}} \phi \left( \frac{\mu}{\sqrt{1 + \sigma^2}} \right) .\label{eq:deriv_gelu}
\end{align}

Equations~\eqref{eq:mean_heaviside}, \eqref{eq:mean_relu}, and~\eqref{eq:mean_gelu} allow us to compute $\boldsymbol{\mu_z}$ in terms of the moments of $\mathbf{y}$, and with their derivatives we can apply Theorem~\ref{thm:cov} to compute $\mathbf{\Sigma_z}$. Equations~\eqref{eq:mean_heaviside} and~\eqref{eq:mean_relu} can be derived with direct integration~\citep{gradshteyn1994}, and in the context of neural networks solutions first appear in~\cite{frey_variational_1999}. We derive Equation~\eqref{eq:mean_gelu}, and give formulae to compute the $k$-th derivative of Equations~\eqref{eq:mean_heaviside}, \eqref{eq:mean_relu}, and~\eqref{eq:mean_gelu}, respectively, in Appendix~\ref{sec:appendix_moments}.

We again note that Theorem~\ref{thm:cov} assumes the activation mean and its derivatives are well-defined, and applies only to activation functions that take the element-wise independent and identical form given in Definition~\ref{thm:def}. Some activation functions, like the sigmoid, don't have a closed-form mean, but accurate approximations exist~\citep{daunizeau_semi-analytical_2017}. We discuss the sigmoid function in detail in Appendix~\ref{sec:appendix_sigmoid}. Some common output layers like the softmax function don't meet the element-wise independent and identical requirement, but also do not present a major obstacle since moments can still be computed for output logits. However, extensions are needed to admit other functions frequently used at intermediate layers, like pooling and batch normalization. \cite{shriver2022} shows that such functions can be refactored into sequences of linear and ReLU operations, to which the moment propagation techniques above can be applied. (Convolution, a linear operation, can be covered straightforwardly by Equations~\eqref{eq:mean_y} and~\eqref{eq:cov_y}.) For the case of pooling, \cite{gast2018lightweight} approximate the mean and variance with closed-form approximations. As an alternative, \cite{springenberg2015striving} show that pooling layers can be replaced with strided convolutions without loss of inference accuracy.

\subsection{Propagation Across Neural Networks}

Since $\mathbf{x}^{(\ell + 1)} = \mathbf{z}^{(\ell)}$, the moment propagation equations in Section~\ref{sec:prop_activations} yield the following algorithm, versions of which appear in~\cite{frey_variational_1999}, \cite{gast2018lightweight}, and \cite{wu2019}:
\begin{algorithm}
\caption{Neural network moment propagation}\label{alg:moment_prop}
\begin{algorithmic}[1]
\STATE Get input mean $\boldsymbol{\mu}_\mathbf{x}$ and covariance $\mathbf{\Sigma}_\mathbf{x}$
\FOR{each network layer}
    \STATE $\boldsymbol{\mu}_\mathbf{y}, \mathbf{\Sigma}_\mathbf{y} \leftarrow \mathbf{W}\boldsymbol{\mu}_\mathbf{x} + \mathbf{b}, \mathbf{W}\mathbf{\Sigma}_\mathbf{x}\mathbf{W}^\top$
    \STATE Compute $\boldsymbol{\mu}_\mathbf{z}, \mathbf{\Sigma}_\mathbf{z}$ from $\mathbf{y} \sim \mathcal{N}(\boldsymbol{\mu}_\mathbf{y}, \mathbf{\Sigma}_\mathbf{y})$
    \STATE $\boldsymbol{\mu}_\mathbf{x}, \mathbf{\Sigma}_\mathbf{x} \leftarrow \boldsymbol{\mu}_\mathbf{z}, \mathbf{\Sigma}_\mathbf{z}$
\ENDFOR
\end{algorithmic}
\end{algorithm}

Algorithm~\ref{alg:moment_prop} lets us analytically estimate the output mean and covariance of a neural network in terms of the input. Output mean and covariance can be used to analyze the robustness of a neural network to noise, adversarial attacks, and other perturbations, as demonstrated by~\cite{bibi2018}.

\subsection{Training Neural Networks}

With moment propagation, we can train neural networks that achieve high inference accuracy, as with standard neural network training, and represent output uncertainty due to input uncertainty~\citep{gast2018lightweight, loquercio2020uncertainty}. This uncertainty is sometimes defined as \emph{aleatoric} uncertainty---uncertainty inherent to the underlying process or system. By altering standard output layers and loss functions---e.g., for regression problems, we can replace mean-squared-error loss with a log-likelihood loss to capture both mean and variance of the output---we can use moment propagation in a supervised training regime to characterize the aleatoric uncertainty of the model's predictions.  In their probabilistic training, \cite{gast2018lightweight} and \cite{loquercio2020uncertainty} only model variance for each neuron, i.e., the diagonal of the full activation covariance matrix $\mathbf{\Sigma_z}$. Using Algorithm~\ref{alg:moment_prop} will capture the interactions between a layer's activation random variables, at the cost of added computational complexity. For a fuller discussion of training probabilistic networks of this form, we refer readers to \cite{gast2018lightweight}.

We can also use moment propagation in the training of fully Bayesian neural networks \citep{wu2019}. As discussed in Section~\ref{sec:related_bnn}, BNNs represent both aleatoric uncertainty and \emph{epistemic} uncertainty---uncertainty in the model, which can be resolved with additional training data---by attempting to characterize the entire predictive distribution instead of learning a single estimate of network parameters. For a dataset of $D$ input-output pairs $\mathcal{D} = \left\{ (\mathbf{x}^{(i)}, \mathbf{z}^{(i)}) \right\}_{i=1}^D$, and a model parameterized by $\boldsymbol{\theta}$, the predictive distribution is \citep{mackay2002book}:
\begin{equation}
p(\mathbf{z}|\mathbf{x},\mathcal{D}) = \int_{\boldsymbol{\theta}} p(\mathbf{z}|\mathbf{x}, \boldsymbol{\theta}) p(\boldsymbol{\theta}|\mathcal{D}) d\boldsymbol{\theta}.
\end{equation}
Because computing the posterior distribution is intractable for modern neural networks, approximative methods like Monte Carlo variational inference (MCVI) are typically used \citep{graves2011variational, kingma2015reparam, wu2019}. MCVI methods minimize the Kullback-Leibler (KL) divergence $D_{KL}$ between a parameterized distribution $q_{\boldsymbol{\phi}}(\boldsymbol{\theta})$ and the true posterior $p(\boldsymbol{\theta}|\mathcal{D})$, where $\boldsymbol{\phi}$ represents the parameters of the variational distribution. Because the true posterior is unknown, MCVI methods instead maximize a surrogate objective, the evidence lower bound (ELBO):
\begin{equation}\label{eq:elbo}
L(\boldsymbol{\phi}) = \mathbb{E}_{q_{\boldsymbol{\phi}}(\boldsymbol{\theta})} \left[ \log p(\mathcal{D}|\boldsymbol{\theta}) \right] - D_{KL} \left( q_{\boldsymbol{\phi}}(\boldsymbol{\theta}) \| p(\boldsymbol{\theta}) \right).
\end{equation}
By choosing a tractable prior $p(\boldsymbol{\theta})$ and variational distribution $q_{\boldsymbol{\phi}}(\boldsymbol{\theta})$ such as factorized (i.e., diagonal covariance matrix) Gaussians, one can iteratively optimize the ELBO with gradient ascent, similar to the process of standard neural network training. However, the first term in Equation~\eqref{eq:elbo}, sometimes called the reconstruction term, can't be computed analytically for modern neural networks, and Monte Carlo sampling is used instead.

\cite{wu2019} present an alternative, sample-free variational inference method, deterministic variational inference (DVI), which they use to estimate Equation~\eqref{eq:elbo} analytically. DVI uses a version of Algorithm~\ref{alg:moment_prop} to approximate MCVI in the limit of infinite samples, $s \rightarrow \infty$. To propagate through a single network layer with an $n \times n$ transform, MCVI has an asymptotic  runtime complexity of $O(sn^2)$, and DVI $O(n^3)$. For an affine transformation $\mathbf{y} = \mathbf{W}\mathbf{x} + \mathbf{b}$, \cite{wu2019} show that, under assumptions of independence between $\mathbf{x}, \mathbf{W}, \mathbf{b}$,
\begin{align}
    \text{Cov}(y_i, y_k) &= \text{Cov}(b_i, b_k) + \nonumber \\
    \sum_j \sum_\ell & \mathbb{E}[x_j x_\ell] \text{Cov}(W_{ij}, W_{k\ell}) + \nonumber \\
    &\mathbb{E}[W_{ij}]\mathbb{E}[W_{k\ell}] \text{Cov}(x_j, x_\ell). \label{eq:dvi_cov}
\end{align}
Because $\mathbf{W}$ and $\mathbf{b}$ are often chosen to be factorized Gaussians for tractability, the only difficult terms to model in Equation~\eqref{eq:dvi_cov} are those involving $\text{Cov}(x_j, x_\ell)$. ($\text{Cov}(W_{ij}, W_{k\ell}) = 0$ for $j \neq \ell$.) To compute $\text{Cov}(x_j, x_\ell)$ in Equation~\eqref{eq:dvi_cov}, \cite{wu2019} rely on activation-specific approximations of Heaviside and ReLU. We can instead apply Theorem~\ref{thm:cov} at no additional asymptotic runtime cost, and extend DVI to any supported activation function.

\section{EXPERIMENTS}\label{sec:experiments}

In experiments, we numerically assess the accuracy of Theorem~\ref{thm:cov} in Section~\ref{sec:error_analysis}, apply it to the analysis of trained neural networks in Section~\ref{sec:trained_experiments}, and apply it to training Bayesian neural networks in Section~\ref{sec:bnn_experiments}. Error analysis in Section~\ref{sec:error_analysis} was performed on an Apple M1 Max, and subsequent experiments were performed on an Nvidia Titan RTX GPU.

\subsection{Error Analysis}\label{sec:error_analysis}

\begin{figure}[t]
\centering
\includegraphics[height=2in]{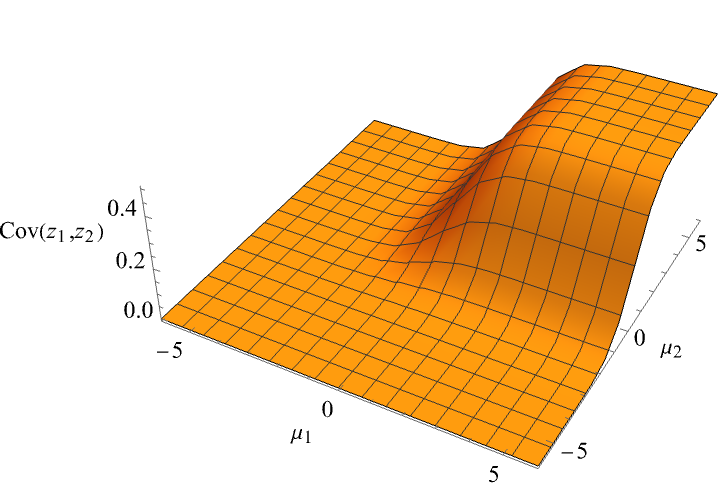}
\caption{The covariance between a bivariate Gaussian passed through a ReLU, calculated numerically, plotted over input means $\mu_1$ and $\mu_2$, with fixed input parameters $\sigma_1 = 1, \sigma_2 = 1, \rho=0.5$.}
\label{fig:covar}
\end{figure}

\begin{figure}[t]
    \centering
    \begin{subfigure}[b]{0.23\textwidth}
        \centering
        \includegraphics[height=4cm]{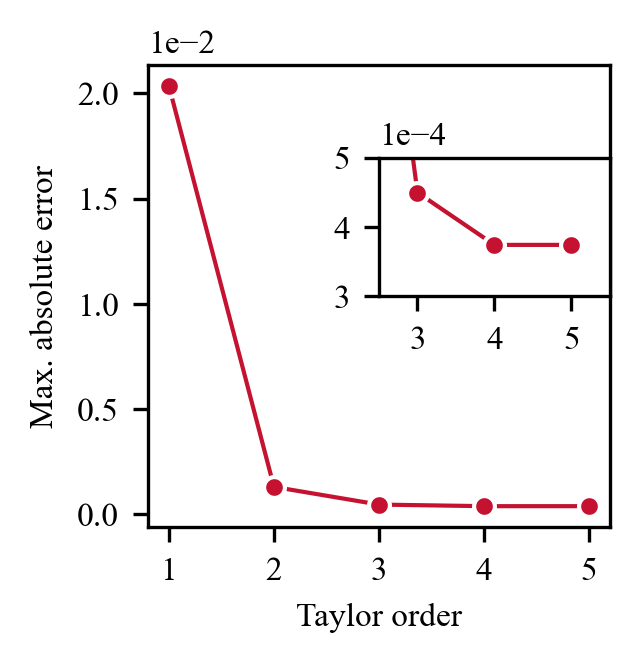}
        \caption{}
    \end{subfigure}
    \begin{subfigure}[b]{0.23\textwidth}
        \centering
        \includegraphics[height=4cm]{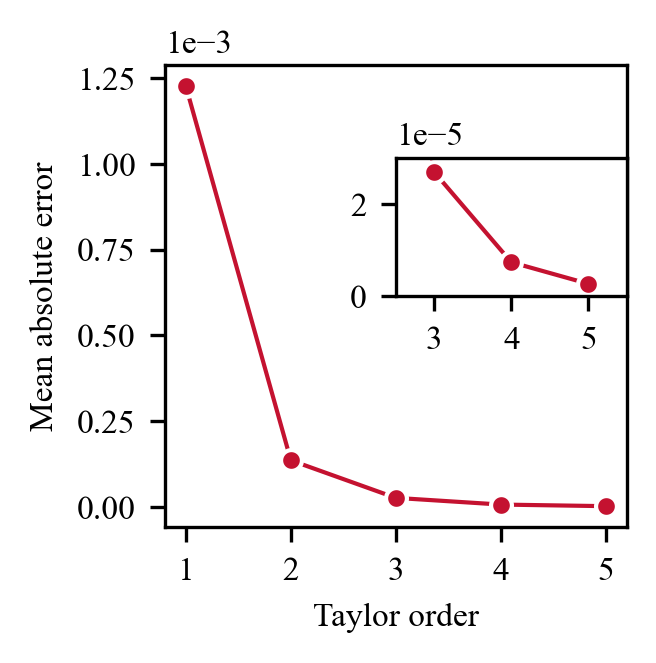}
        \caption{}
    \end{subfigure}
    \caption{The (a) maximum and (b) mean absolute error of Theorem~\ref{thm:cov} applied to a ReLU by Taylor order. Error is determined by comparing against the numerical calculation depicted in Figure~\ref{fig:covar}, over $\mu_1,\mu_2 \in \left[ -5, 5 \right]$.}
    \label{fig:taylor_order}
\end{figure}

\begin{figure}[t]
     \centering
     \begin{subfigure}[b]{0.23\textwidth}
         \centering
         \includegraphics[width=\textwidth]{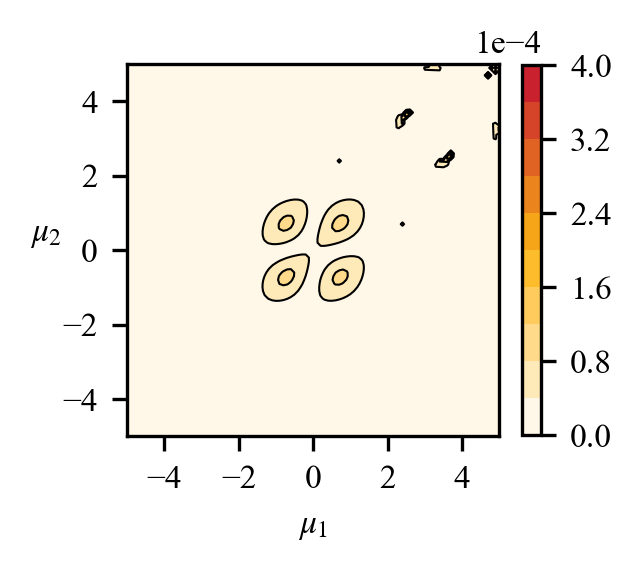}
         \caption{}
         \label{fig:sub_relu_err}
     \end{subfigure}
     \begin{subfigure}[b]{0.23\textwidth}
         \centering
         \includegraphics[width=\textwidth]{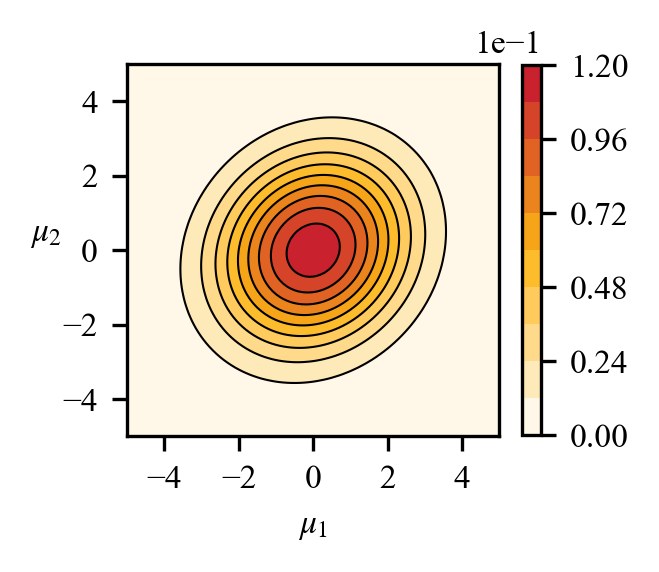}
         \caption{}
         \label{fig:sub_wu_err}
     \end{subfigure}
     \caption{Absolute error for (a) a fourth-order expansion of Theorem~\ref{thm:cov} applied to a ReLU and (b) the ReLU approximation proposed by~\cite{wu2019}, plotted over input means $\mu_1$ and $\mu_2$. Error is determined by comparing against the numerical calculation depicted in Figure~\ref{fig:covar}.}
     \label{fig:error_comparison}
\end{figure}

\begin{figure}
     \centering
     \begin{subfigure}[b]{0.23\textwidth}
         \centering
         \includegraphics[width=\textwidth]{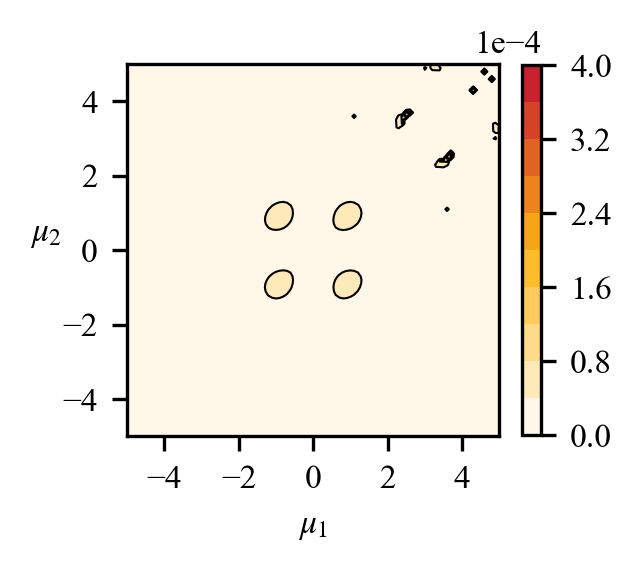}
         \caption{}
         \label{fig:gelu_error}
     \end{subfigure}
     \begin{subfigure}[b]{0.23\textwidth}
         \centering
         \includegraphics[width=\textwidth]{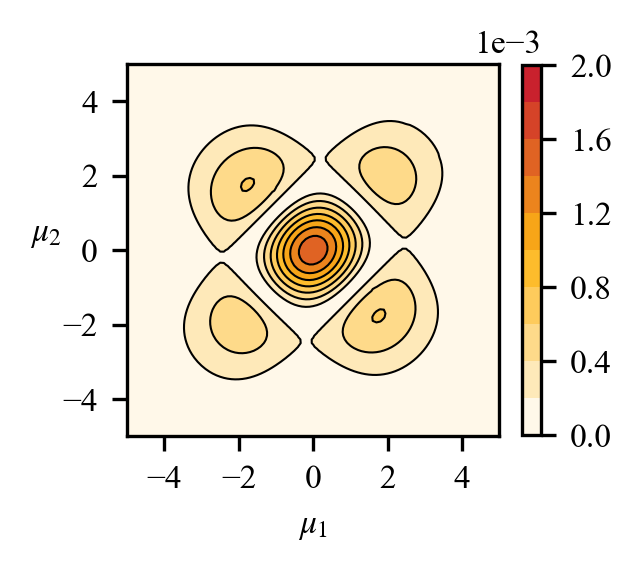}
         \caption{}
         \label{fig:sigmoid_err}
     \end{subfigure}
     \caption{Absolute error of a fourth-order expansion of Theorem~\ref{thm:cov} applied to (a) a GELU and (b) an approximated sigmoid~\citep{daunizeau_semi-analytical_2017}, plotted over input means $\mu_1$ and $\mu_2$ and using the fixed input parameters of Figure~\ref{fig:covar}.}
     \label{fig:error_other}
\end{figure}

Figure~\ref{fig:covar} shows the covariance of a bivariate Gaussian passed through a ReLU, calculated numerically. We can compare this numerical calculation to the analytic solution of Theorem~\ref{thm:cov}. Figure~\ref{fig:taylor_order} shows the error by Taylor order, and Figure~\ref{fig:error_comparison} compares the the error with the approximation proposed by \cite{wu2019}. The \cite{wu2019} approximation has a maximum absolute error of $1.160 \times 10^{-1}$. A first-order approximation of Theorem~\ref{thm:cov} has a maximum absolute error of $2.034 \times 10^{-2}$, and a fourth-order approximation $3.740 \times 10^{-4}$. We can similarly analyze the error of Theorem~\ref{thm:cov} applied to other activation functions for which covariance approximations don't appear in previous work, like the GELU or the approximated sigmoid~\citep{daunizeau_semi-analytical_2017}, both shown in Figure~\ref{fig:error_other}. The GELU and sigmoid are discussed more fully in Appendix~\ref{sec:appendix_moments}.

\subsection{Characterizing Trained Neural Networks}\label{sec:trained_experiments}

\begin{table*}[ht]
\centering
\caption{Tightness measures for synthetic data and networks of varying type and depth, comparing the proposed moment propagation method to PL-DNN~\citep{bibi2018}. We report the mean and standard deviation of the ratio of mean estimates $\Qoppa{\hat{\mu}}=\hat{\mu}_{MC}/\hat{\mu}_{A}$ and ratio of variance estimates $\Qoppa{\hat{\sigma}^2=\hat{\sigma}^2_{MC}/\hat{\sigma}^2_{A}}$ between the Monte Carlo and analytic methods, calculated over 200 instances. Ideally the ratio approaches $1 \pm 0$.}
\label{tab:tightness_synthetic}
\begin{center}
\begin{tabular}{rrrrr} \toprule
\ & \multicolumn{2}{c}{Proposed} & \multicolumn{2}{c}{PL-DNN} \\
\cmidrule(lr){2-3} \cmidrule(lr){4-5}
Network & \multicolumn{1}{c}{${\Qoppa}{\hat{\mu}}$} & \multicolumn{1}{c}{$\Qoppa{\hat{\sigma}^2}$} & \multicolumn{1}{c}{${\Qoppa}{\hat{\mu}}$} & \multicolumn{1}{c}{$\Qoppa{\hat{\sigma}^2}$}
\\ \midrule
FC-4 & $1.000 \pm 0.014$ & $1.010 \pm 0.011$ & $0.999 \pm 0.643$ & $1.155 \pm 0.311$ \\
FC-8 & $1.000 \pm 0.001$ & $1.016 \pm 0.017$ & $1.012 \pm 0.169$ & $1.067 \pm 0.419$ \\
CNN-4 & $1.001 \pm 0.021$ & $1.007 \pm 0.006$ & $0.997 \pm 0.168$ & $1.222 \pm 0.240$ \\
CNN-8 & $1.000 \pm 0.000$ & $1.009 \pm 0.007$ & $1.001 \pm 0.026$ & $1.184 \pm 0.347$ \\
\bottomrule
\end{tabular}
\end{center}
\end{table*}

\begin{table*}[ht]
\centering
\caption{Tightness measures for a convolutional network trained on MNIST data, comparing the proposed moment propagation method to PL-DNN~\citep{bibi2018}. For each of the 10 class logits, we report the mean and standard deviation of the ratio of mean estimates $\Qoppa{\hat{\mu}}=\hat{\mu}_{MC}/\hat{\mu}_{A}$ and ratio of variance estimates $\Qoppa{\hat{\sigma}^2}=\hat{\sigma}^2_{MC}/\hat{\sigma}^2_{A}$ between the Monte Carlo and analytic methods, calculated over 200 instances.}
\label{tab:tightness_mnist}
\begin{center}
\begin{tabular}{rrrrr} \toprule
\ & \multicolumn{2}{c}{Proposed} & \multicolumn{2}{c}{PL-DNN} \\
\cmidrule(lr){2-3} \cmidrule(lr){4-5}
Logit & \multicolumn{1}{c}{${\Qoppa}{\hat{\mu}}$} & \multicolumn{1}{c}{$\Qoppa{\hat{\sigma}^2}$} & \multicolumn{1}{c}{${\Qoppa}{\hat{\mu}}$} & \multicolumn{1}{c}{$\Qoppa{\hat{\sigma}^2}$}
\\ \midrule
$0$ & $1.000 \pm 0.000$ & $1.001 \pm 0.007$ & $1.016 \pm 0.102$ & $1.422 \pm 0.229$ \\
$1$ & $1.001 \pm 0.001$ & $1.004 \pm 0.005$ & $0.971 \pm 0.185$ & $1.553 \pm 0.254$ \\
$2$ & $1.001 \pm 0.001$ & $1.009 \pm 0.008$ & $0.998 \pm 0.267$ & $1.531 \pm 0.222$ \\
$3$ & $0.999 \pm 0.007$ & $1.009 \pm 0.008$ & $1.018 \pm 0.410$ & $1.512 \pm 0.252$ \\
$4$ & $1.000 \pm 0.001$ & $1.005 \pm 0.006$ & $0.998 \pm 0.141$ & $1.383 \pm 0.244$ \\
$5$ & $1.001 \pm 0.015$ & $1.006 \pm 0.006$ & $0.980 \pm 0.696$ & $1.733 \pm 0.267$ \\
$6$ & $1.001 \pm 0.014$ & $1.004 \pm 0.007$ & $0.998 \pm 0.096$ & $1.382 \pm 0.217$ \\
$7$ & $1.000 \pm 0.003$ & $1.009 \pm 0.009$ & $0.975 \pm 0.178$ & $1.500 \pm 0.161$ \\
$8$ & $1.000 \pm 0.002$ & $1.005 \pm 0.006$ & $1.068 \pm 0.684$ & $1.453 \pm 0.205$ \\
$9$ & $1.000 \pm 0.002$ & $1.004 \pm 0.005$ & $0.993 \pm 0.043$ & $1.557 \pm 0.297$ \\
\bottomrule
\end{tabular}
\end{center}
\end{table*}

\begin{table*}[ht]
\caption{Test log-likelihoods for UCI regression datasets. We compare the proposed method to DVI~\citep{wu2019} and MCVI, keeping the training routine and model architecture otherwise identical. For each dataset we report the mean and standard deviation for 20 randomized data splits. A higher test log-likelihood is better, and best results are in boldface.}
\label{tab:uci}
\begin{center}
\begin{tabular}{rrrrrr} \toprule
\multicolumn{3}{c}{Dataset} & \multicolumn{3}{c}{Method} \\
\cmidrule(lr){1-3} \cmidrule(lr){4-6}
Name & Samples & Features & Proposed & DVI & MCVI
\\ \midrule
\texttt{boston} & 506 & 13 & $\boldsymbol{-2.31 \pm 0.19}$ & $-2.33 \pm 0.19$ & $-2.40 \pm 0.23$ \\
\texttt{concrete} & 1,030 & 8 & $\boldsymbol{-2.98 \pm 0.11}$ & $-2.99 \pm 0.11$ & $-3.04 \pm 0.10$ \\
\texttt{energy} & 768 & 8 & $\boldsymbol{-1.27 \pm 0.21}$ & $-1.30 \pm 0.22$ & $-1.40 \pm 0.18$ \\
\texttt{kin8nm} & 8,192 & 8 & $1.11 \pm 0.03$ & $1.11 \pm 0.03$ & $\boldsymbol{1.20 \pm 0.03}$ \\
\texttt{naval} & 11,934 & 16 & $5.81 \pm 0.16$ & $5.73 \pm 0.22$ & $\boldsymbol{5.92 \pm 0.24}$ \\
\texttt{power} & 9,568 & 4 & $\boldsymbol{-2.82 \pm 0.04}$ & $\boldsymbol{-2.82 \pm 0.04}$ & $\boldsymbol{-2.82 \pm 0.04}$ \\
\texttt{protein} & 45,730 & 9 & $\boldsymbol{-2.97 \pm 0.01}$ & $\boldsymbol{-2.97 \pm 0.02}$ & $-3.05 \pm 0.02$ \\
\texttt{wine} & 1,588 & 11 & $\boldsymbol{-0.91 \pm 0.07}$ & $-0.92 \pm 0.07$ & $-0.94 \pm 0.07$ \\
\texttt{yacht} & 308 & 6 & $-0.25 \pm 0.20$ & $\boldsymbol{-0.23 \pm 0.22}$ & $-0.53 \pm 0.19$ \\
\bottomrule
\end{tabular}
\end{center}
\end{table*}

We study how well moment propagation can characterize the input-output distribution of a trained network by comparing to Monte Carlo estimation.

Following the experiments in \cite{bibi2018}, we first test our method on synthetic data and networks. We construct both fully connected and convolutional networks of varying depth, with ReLU activations throughout and i.i.d. network parameters generated with Kaiming initialization \citep{kaiming2015}. Input data are $\mathbb{R}^{100}$ for the fully connected networks and $\mathbb{R}^{20 \times 20}$ for the convolutional networks, and input noise is represented with a multivariate Gaussian distribution. In the fully connected networks, each hidden layer is 100 units. In the convolutional networks, each convolutional layer is 10 channels with $3\times 3$ filter size and ``same'' padding to preserve feature map dimensions. The mean is drawn from a standard normal Gaussian, and the covariance is based on randomly drawn eigenvalues such that the maximum variance is $1.0$. All networks output a single scalar value.

In each experiment, we generate a random mean and covariance for the input distribution, from which we draw $7.5 \times 10^4$ Monte Carlo samples to pass through the network and form an estimate of the output mean and variance. We then analytically propagate the input mean and covariance using both our proposed method and the PL-DNN technique proposed in \cite{bibi2018}. Each experiment is repeated 200 times.

Table~\ref{tab:tightness_synthetic} shows tightness measures for fully connected and convolutional networks with 4 and 8 layers (denoted as FC-4, FC-8, CNN-4, and CNN-8). The tightness of the analytic methods is measured by taking the ratio of the Monte Carlo estimates of mean and variance to the respective analytic estimates. (We use $\Qoppa \hat{\mu}$ to represent the ratio of estimates as one might use $\Delta \hat{\mu}$ to represent the difference of estimates.) We see that the proposed method achieves high estimation accuracy for both mean and variance, with a better average tightness and lower standard deviation than PL-DNN. We conjecture that this is because, while both methods approximate unknown distributions as Gaussian, PL-DNN approximates the known structure of the network whereas our method preserves it.

The weights of a trained neural network are not random, of course, and training may induce correlations between network parameters. As in \cite{bibi2018}, we also compare methods using a network trained on MNIST data \citep{mnist2010} with a convolutional architecture based on LeNet \citep{lecun1999}. Because our proposed technique doesn't admit pooling layers without extension, we replace pooling layers with strided convolutions \citep{springenberg2015striving}. The network has two convolutional layers with ReLU activations, each followed by a strided convolution in place of a pooling layer, then two fully connected layers. The network input is $\mathbb{R}^{28 \times 28}$ corresponding to an image, and the output is a set of 10 logits corresponding to the 10 possible classes. For each experiment, we represent the input as zero-mean Gaussian noise with a randomly generated covariance matrix added to an image $\mathbf{M}$ randomly selected from the MNIST test set. As in the synthetic experiments, we generate $7.5 \times 10^4$ Monte Carlo samples to estimate output mean and variance, analytically propagate input mean and covariance using both the proposed method and PL-DNN, and repeat each experiment 200 times.

Table~\ref{tab:tightness_mnist} shows tightness measures for each output logit, computed in the same manner as in the experiments on synthetic data. Similarly, the proposed method achieves a consistently high estimation accuracy for both mean and variance, with a better average tightness and lower standard deviation than PL-DNN.

\subsection{Training Bayesian Neural Networks}\label{sec:bnn_experiments}

In our training experiments, we train small BNNs on UCI regression datasets~\citep{uci}. We compare different implementations of variational inference: baseline Monte Carlo variational inference (MCVI), deterministic variational inference (DVI) as proposed by \cite{wu2019}, and our version of DVI which uses Theorem~\ref{thm:cov} to calculate ReLU covariance more accurately. We manually tune the learning rate, but otherwise use the same training routine, hyperparameters, and model architecture across the different datasets and implementations. The model is heteroskedastic (i.e., it outputs both a mean and variance) and consists of a single hidden layer of 50 units with fixed priors and ReLU activations (except for the \texttt{protein} dataset, for which we use 100 units). Following \cite{wu2019}, we choose a factorized Gaussian for our variational distribution (i.e., the covariance matrix of trainable parameters is diagonal) to reduce computational complexity. We use the AdamW optimizer \citep{loshchilov2018decoupled}, the batch size is fixed at 10 and training is run for 50 epochs (except for the larger \texttt{protein} dataset, which is run for 25). For MCVI, the number of Monte Carlo samples is fixed at 10. Each dataset is randomly split into 90\% training data and 10\% test data, and each training experiment is repeated over 20 randomized data splits.

In Table~\ref{tab:uci} we report the mean and standard deviation of the best test log-likelihood---the reconstruction term in Equation~\eqref{eq:elbo}---for each dataset and method. On average, the more accurate covariance calculation from Theorem~\ref{thm:cov} modestly improves BNN training, but this effect is not uniform, and in some cases performance drops. We conjecture that in such cases the ReLU covariance approximation used by \cite{wu2019} is sufficiently accurate, or that the factorized Gaussian assumption curbs the impact of accurate covariance calculations and other aspects of the learning problem predominate. We also note that the DVI and MCVI results do not exactly match those reported in \cite{wu2019}, as we run the training experiments with our own BNN implementation; we attribute discrepancies to differences in hyperparameter selection.

\section{CONCLUSION}

We introduced a general, analytic solution to compute activation covariance and illustrated how this can propagate moments across layers in neural networks. Our experiments confirm the validity of this approach. Our moment propagation method can be applied to characterizing the input-output distributions of trained neural networks and to training probabilistic neural networks that model predictive uncertainty, both of which offer a path to neural networks that are quantifiably reliable and robust.

Prominent research questions remain for successful uncertainty quantification in neural networks. Investigating the application of Theorem~\ref{thm:cov} to Laplace approximation~\citep{daxberger_laplace_2021} may be fruitful. Because computing a covariance matrix is of course $\mathcal{O}(n^2)$ in the number of dimensions $n$, covariance-based methods still face significant scaling challenges. Also germane to this work are studying more thoroughly the effect of activation covariance, and formulating a theory for the applicability of the Gaussian approximation.

\subsubsection*{Acknowledgements}
We thank Anqi Wu and Adel Bibi for their generous correspondence; Eric Heim, Shannon Gallagher, and David Shriver for their comments; and J. Nelson Wright for his insight into Fourier analysis. This material is based upon work funded in part by the Department of Defense under Contract No. FA8702-15-D-0002 with Carnegie Mellon University for the operation of the Software Engineering Institute, a federally funded research and development center; in part by JST, PRESTO Grant Number JPMJPR2136, Japan; in part by the Department of the Navy, Office of Naval Research, grant number N00014-23-1-2252; in part by Mobility21 National University Transportation Center, which is sponsored by the Department of Transportation; and in part by NSF grant CCF 2327905.

[DISTRIBUTION STATEMENT A] This material has been approved for public release and unlimited distribution.  Please see Copyright notice for non-US Government use and distribution. DM24-0157.

\bibliography{bibliography}

\appendix
\onecolumn
\section{COVARIANCE PROPAGATION, PROOF}\label{sec:appendix_cov}

We recall Theorem~\ref{thm:cov} and provide a complete derivation below.
\begin{reptheorem}{thm:cov}
Suppose $\mathbf{y} \in \mathbb{R}^{n}$ is a multivariate Gaussian random vector, $\mathbf{y} \sim \mathcal{N}(\boldsymbol{\mu}, \mathbf{\Sigma)}$, and $\mathbf{z} = \mathbf{g}(\mathbf{y})$ is an element-wise independent and identical function of $\mathbf{y}$. If the mean and variance of any input element $y_i$ is $(\mu_i, \sigma^2_i)$, and any two input elements $y_i, y_j$ are related by Pearson correlation coefficient $\rho_{ij} = \mathbb{E}[(y_i - \mu_i)(y_j - \mu_j)]/(\sigma_i \sigma_j)$, then the covariance between output elements $z_i$ and $z_j$ is
\begin{equation*}
    \Cov(z_i, z_j) = \sum^{\infty }_{k=1} \frac{\rho^{k}_{ij}}{k!}
    \biggl(
  	\sigma_{i}^{k}\ \frac{\partial^{k} \mathbb{E}[z_i] }{\partial \mu^{k}_{i}}
    \biggr)
    \biggl(
  	\sigma_{j}^{k}\ \frac{\partial^{k} \mathbb{E}[z_j] }{\partial \mu^{k}_{j}}
    \biggr).
\end{equation*}
\end{reptheorem}
\begin{proof}
With Gaussian random variables, the integrals of interests are (omitting subscripts unless necessary):
\begin{align}
\mathbb{E}[z] &= \frac{1}{\sqrt{2 \pi \sigma^2}} \int_{-\infty}^{\infty} g(y) \exp\left\{ \frac{-(y-\mu)^2}{2 \sigma^2} \right\} dy \label{eq:moment_integral} \\
\mathbb{E}[z^2] &= \frac{1}{\sqrt{2 \pi \sigma^2}} \int_{-\infty}^{\infty} g(y)^2 \exp\left\{ \frac{-(y-\mu)^2}{2 \sigma^2} \right\} dy \label{eq:moment2_integral} \\
\mathbb{E}[z_i z_j] &= \frac{1}{2\pi \sigma_{i} \sigma_{j} \sqrt{1-\rho^{2} } } \int_{-\infty}^{\infty} \int_{-\infty}^{\infty}   g(y_{i}) g(y_{j}) \nonumber \\ &\exp \left\{
-\frac{1}{2\left( 1-\rho^{2} \right)  } \left[ \frac{\left( y_{i}-\mu_{i}\right)^{2}  }{\sigma^{2}_{i} } -\frac{2\rho \left( y_{i}-\mu_{i}\right)  \left( y_{j}-\mu_{j}\right)  }{\sigma_{i} \sigma_{j} } +\frac{\left( y_{j}-\mu_{j}\right)^{2}  }{\sigma^{2}_{j} } \right] \right\}  dy_{i} dy_{j} \label{eq:momentx_integral}.
\end{align}

We recognize the integral in Equation~\eqref{eq:moment_integral} as a convolution in $\mu$, and make use of the Fourier transform convolution property. Let $G(\xi) = \mathfrak{F}\left\{ g(\mu) \right\}$, the Fourier transform of $g(\mu)$. The one-dimensional Fourier transforms of interest are:
\begin{align*}
g(\mu) &\xrightarrow{\mathfrak{F}} G(\xi) \\
g(\mu)^2 &\xrightarrow{\mathfrak{F}} \left( G*G \right)(\xi) \\
\frac{1}{\sqrt{2\pi \sigma^2} } \exp\left\{ -\frac{\mu^{2}}{\sigma^{2} } \right\}  &\xrightarrow{\mathfrak{F}} \exp\left\{ -2  \pi^{2} \xi^{2} \sigma^{2} \right\}.
\end{align*}
Let $ H(\xi) = \left( G*G \right)(\xi)$. We can express Equations~\eqref{eq:moment_integral} and~\eqref{eq:moment2_integral} as inverse Fourier transforms:
\begin{align}
    \mathbb{E}[z] &= \mathfrak{F}^{-1} \left\{ G(\xi) \exp \{ -2\pi^2 \xi^2 \sigma^2 \} \right\} \label{eq:moment_invF} \\
    \mathbb{E}[z^2] &= \mathfrak{F}^{-1} \left\{ H(\xi) \exp \{ -2\pi^2 \xi^2 \sigma^2 \} \right\}. \label{eq:moment2_invF}
\end{align}

Similarly, we can treat Equation~\eqref{eq:momentx_integral} as a two-dimensional convolution and exploit this to determine $\mathbb{E}[z_i z_j]$. With the two-dimensional Fourier transform of a bivariate Gaussian
\begin{equation*}
\frac{1}{2\pi \sigma_{i} \sigma_{j} \sqrt{1-\rho^{2} } } \exp\left\{ -\frac{1}{2\left( 1-\rho^{2} \right)  } \left[ \frac{\mu^{2}_{i}}{\sigma^{2}_{i} } -\frac{2\rho \mu_{i}\mu_{j}}{\sigma_{i} \sigma_{j} } +\frac{\mu^{2}_{j}}{\sigma^{2}_{j} } \right]  \right\} \xrightarrow{\mathfrak{F}} \exp\left\{ -2  \pi^{2} \left[ \xi^{2}_{i} \sigma^{2}_{i} +2\rho \xi_{i} \xi_{j} \sigma_{i} \sigma_{j} +\xi^{2}_{j} \sigma^{2}_{j} \right]  \right\}
\end{equation*}
we can express Equation~\eqref{eq:momentx_integral} as an inverse Fourier transform:
\begin{equation}
\mathbb{E}[z_i z_j] = \mathfrak{F}^{-1} \Bigl\{ G(\xi_{i} )G(\xi_{j} )
\exp\bigl\{ -2\pi^{2} [\xi^{2}_{i} \sigma^{2}_{i} +2\rho \xi_{i} \xi_{j} \sigma_{i} \sigma_{j} +\xi^{2}_{j} \sigma^{2}_{j} ]\bigr\}  \Bigr\}. \label{eq:momentx_invF}
\end{equation}

Using Equations~\eqref{eq:moment_invF} and~\eqref{eq:momentx_invF}, we can then express the covariance $\mathrm{Cov}(z_i, z_j) = \mathbb{E}[z_i z_j] - \mathbb{E}[z_i]\mathbb{E}[z_j]$ as
\begin{equation*}
\mathrm{Cov}(z_i, z_j) = \mathfrak{F}^{-1} \Bigl\{ G\left( \xi_{i} \right)  G\left( \xi_{j} \right)  \exp\left\{ -2\pi^{2} \xi^{2}_{i} \sigma^{2}_{i} \right\}  
\left[ \exp\left\{ -4\pi^{2}\rho \xi_{i} \xi_{j} \sigma_{i} \sigma_{j} \right\}  -1\right]  \exp\left\{ -2\pi^{2} \xi^{2}_{j} \sigma^{2}_{j} \right\}  \Bigr\}.
\end{equation*}
To solve this inverse Fourier transform, we can express $ \exp\left\{ -4\pi^{2} \rho \xi_{i} \xi_{j} \sigma_{i} \sigma_{j} \right\}  -1$ as a Taylor series in $\rho$:
\begin{equation*}
\exp\left\{ -4\pi^{2} \rho \xi_{i} \xi_{j} \sigma_{i} \sigma_{j} \right\}  -1 = \sum^{\infty }_{k=1} \frac{\rho^{k} }{k!} \left( -4\pi^{2} \xi_i \xi_j \sigma_{i} \sigma_{j} \right)^{k}.
\end{equation*}
We then move the summation out of the inverse Fourier operation:
\begin{equation*}
\mathrm{Cov}(z_i, z_j) = \sum^{\infty }_{k=1} \frac{\rho^{k} }{k!} \left( -4\pi^{2} \sigma_{i} \sigma_{j} \right)^{k} 
\mathfrak{F}^{-1} \Bigl\{ \xi^{k}_{i} \  \xi^{k}_{j} \  G\left( \xi_{i} \right)  G\left( \xi_{j} \right)
\exp\left\{ -2\pi^{2} \xi^{2}_{i} \sigma^{2}_{i} \right\}  \exp\left\{ -2\pi^{2} \xi^{2}_{j} \sigma^{2}_{j} \right\}  \Bigr\}.  
\end{equation*}
Then, using the separability property of two-dimensional Fourier transforms:
\begin{equation*}
\mathrm{Cov}(z_i, z_j) = \sum^{\infty }_{k=1} \frac{\rho^{k} }{k!} \left( -4\pi^{2} \sigma_{i} \sigma_{j} \right)^{k} 
\mathfrak{F}^{-1} \left\{ \xi^{k}_{i} \  G\left( \xi_{i} \right)   \exp\left\{ -2\pi^{2} \xi^{2}_{i} \sigma^{2}_{i} \right\}    \right\}  \mathfrak{F}^{-1} \left\{   \xi^{k}_{j} \   G\left( \xi_{j} \right)  \exp\left\{ -2\pi^{2} \xi^{2}_{j} \sigma^{2}_{j} \right\}  \right\}.
\end{equation*}
Using the Fourier differentiation property and Equation~\eqref{eq:moment_invF}, we see that
\begin{equation*}
    \xi^{k} \  G\left( \xi \right)   \exp\left\{ -2\pi^{2} \xi^{2} \sigma^{2} \right\} \xrightarrow{\mathfrak{F}^{-1}} \frac{1}{\left( 2\pi \sqrt{-1}\right)^k} \frac{\partial^{k} \mathbb{E}[z]}{\partial \mu^{k}}.
\end{equation*}
Hence, the covariance can be expressed as:
\begin{equation*}
    \Cov(z_i, z_j) = \sum^{\infty }_{k=1} \frac{\rho^{k}_{ij}}{k!}
    \left(
  	\sigma_{i}^{k}\ \frac{\partial^{k} \mathbb{E}[z_i] }{\partial \mu^{k}_{i}}
    \right)
    \left(
  	\sigma_{j}^{k}\ \frac{\partial^{k} \mathbb{E}[z_j] }{\partial \mu^{k}_{j}}
    \right).
\end{equation*}
\end{proof}

\section{MOMENTS OF COMMON ACTIVATION FUNCTIONS}\label{sec:appendix_moments}

\subsection{Heaviside}

The Heaviside step function $u(y)$ is defined as
\begin{equation*}
    u(y) = 
    \begin{cases}
        0 & \text{if } y < 0\\
        1 & \text{if } y \geq 0.
    \end{cases}
\end{equation*}
When the input is a Gaussian random variable $y \sim \mathcal{N}(\mu, \sigma^2)$, the mean and variance of $z = u(y)$, first reported in the context of neural networks by \cite{frey_variational_1999}, are
\begin{align}
    \mathbb{E}[z] &= \Phi \left( \frac{\mu}{\sigma} \right) \label{eq:mean_heaviside_a} \\
    \Var(z) &= \Phi \left( \frac{\mu}{\sigma} \right) \left[1 - \Phi \left( \frac{\mu}{\sigma} \right) \right] \label{eq:var_heaviside_a}
\end{align}

To compute the covariance of the output of a Heaviside step function using Theorem~\ref{thm:cov}, the first few terms of interest are:
\begin{align*}
\sigma \frac{\partial \mathbb{E}[z] }{\partial \mu} &= \phi \left( \frac{\mu}{\sigma} \right) \\
\sigma^2 \frac{\partial ^2 \mathbb{E}[z] }{\partial \mu^2} &= -\left( \frac{\mu}{\sigma} \right) \phi \left( \frac{\mu}{\sigma} \right) \\
\sigma^3 \frac{\partial^3 \mathbb{E}[z] }{\partial \mu^3} &= \left[ \left( \frac{\mu}{\sigma} \right)^2 - 1 \right] \phi \left( \frac{\mu}{\sigma} \right) \\
\sigma^4 \frac{\partial^4 \mathbb{E}[z] }{\partial \mu^4} &= - \left( \frac{\mu}{\sigma} \right) \left[ \left( \frac{\mu}{\sigma} \right)^2 - 3 \right] \phi \left( \frac{\mu}{\sigma} \right) \\
\sigma^5 \frac{\partial^5 \mathbb{E}[z] }{\partial \mu^5} &= \left[ \left( \frac{\mu}{\sigma} \right)^4 - 6\left( \frac{\mu}{\sigma} \right)^2 + 3 \right] \phi \left( \frac{\mu}{\sigma} \right).
\end{align*}

With the probabilist's Hermite polynomial
\begin{equation*}
    \mathrm{He_k}(x) = \frac{\left( -1\right)^{k}  }{\phi \left( x\right)  } \frac{\partial^{k} \phi(x) }{\partial x^{k}} 
\end{equation*}
terms can be expressed as
\begin{equation}\label{eq:heaviside_hermite_a}
\sigma^k \frac{\partial^{k} \mathbb{E}[z] }{\partial \mu^{k}} \  = (-1)^{k-1}\,\mathrm{He_{k-1}}\left(\frac{\mu}{\sigma}\right)\phi \left( \frac{\mu}{\sigma}\right).
\end{equation}

\subsection{Rectified Linear Unit}

The rectified linear unit (ReLU) is defined as
\begin{equation*}
    \text{ReLU}(y) = 
    \begin{cases}
        0 & \text{if } y < 0\\
        y & \text{if } y \geq 0.
    \end{cases}
\end{equation*}
When the input is a Gaussian random variable $y \sim \mathcal{N}(\mu, \sigma^2)$, the mean and variance of $z = \text{ReLU}(y)$, first reported in the context of neural networks by \cite{frey_variational_1999}, are
\begin{align}
    \mathbb{E}[z] &= \mu \Phi \left( \frac{\mu}{\sigma} \right) + \sigma \phi \left( \frac{\mu}{\sigma} \right) \label{eq:mean_relu_a} \\
    \Var(z) &= \left( \mu^2 + \sigma^2 \right) \Phi \left( \frac{\mu}{\sigma} \right) + \mu \sigma \phi\left( \frac{\mu}{\sigma} \right) - \mathbb{E}[z]^2 \label{eq:var_relu_a}
\end{align}

To compute the covariance of the output of a ReLU using Theorem~\ref{thm:cov}, the first few terms of interest are:
\begin{align*}
\sigma \frac{\partial \mathbb{E}[z] }{\partial \mu} &= \sigma \Phi \left( \frac{\mu}{\sigma} \right) \\
\sigma^2 \frac{\partial ^2 \mathbb{E}[z] }{\partial \mu^2} &= \sigma \phi \left( \frac{\mu}{\sigma} \right) \\
\sigma^3 \frac{\partial^3 \mathbb{E}[z] }{\partial \mu^3} &= -\sigma \left( \frac{\mu}{\sigma} \right) \phi \left( \frac{\mu}{\sigma} \right) \\
\sigma^4 \frac{\partial^4 \mathbb{E}[z] }{\partial \mu^4} &= \sigma \left[ \left( \frac{\mu}{\sigma} \right)^2 - 1 \right] \phi \left( \frac{\mu}{\sigma} \right) \\
\sigma^5 \frac{\partial^5 \mathbb{E}[z] }{\partial \mu^5} &= - \sigma \left( \frac{\mu}{\sigma} \right) \left[ \left( \frac{\mu}{\sigma} \right)^2 - 3 \right] \phi \left( \frac{\mu}{\sigma} \right).
\end{align*}

With the probabilist's Hermite polynomial, terms for $k>1$ can be expressed as
\begin{equation}\label{eq:relu_hermite_a}
\sigma^k \frac{\partial^{k} \mathbb{E}[z] }{\partial \mu^{k}} = \sigma(-1)^k\,\mathrm{He_{k-2}}\left(\frac{\mu}{\sigma}\right)\phi \left( \frac{\mu}{\sigma}\right).
\end{equation}

\subsection{Gaussian Error Linear Unit}\label{sec:appendix_gelu}

The Gaussian Error Linear Unit (GELU), first proposed by \cite{hendrycks2016gelu}, is defined as
\begin{equation*}
\text{GELU}(y) = y \Phi(y)
\end{equation*}
When the input is a Gaussian random variable $y \sim \mathcal{N}(\mu, \sigma^2)$, the mean of $z = \text{GELU}(y)$ is
\begin{equation}\label{eq:gelu_mean_a}
\mathbb{E}[z] = \mu \Phi \left( \frac{\mu}{\sqrt{1 + \sigma^2}} \right) + \frac{\sigma^2}{\sqrt{1 + \sigma^2}} \phi \left( \frac{\mu}{\sqrt{1 + \sigma^2}} \right).
\end{equation}

Unlike the Heaviside and ReLU moments, Equation~\eqref{eq:gelu_mean_a} does not appear in prior work. We provide a derivation below.

\begin{proof}
The integral of interest is
\begin{equation}\label{eq:gelu_moment_a}
\mathbb{E}[z] = \frac{1}{\sqrt{2 \pi \sigma^2}} \int_{-\infty}^{\infty} y \Phi(y) \exp\left\{ \frac{-(y-\mu)^2}{2 \sigma^2} \right\} dy.
\end{equation}

With the definition of $\Phi(y)$ we can separate this into two integrals. The first integral
\begin{equation*}
    \frac{1}{2} \frac{1}{\sqrt{2 \pi \sigma^2}} \int_{-\infty}^{\infty} y \exp\left\{ \frac{-(y-\mu)^2}{2 \sigma^2} \right\} dy
\end{equation*}
evaluates to $\frac{\mu}{2}$. The second integral
\begin{equation*}
    \frac{1}{2} \frac{1}{\sqrt{2 \pi \sigma^2}} \int_{-\infty}^{\infty} y \ \text{erf}\left(\frac{y}{\sqrt{2}}\right) \exp\left\{ \frac{-(y-\mu)^2}{2 \sigma^2} \right\} dy
\end{equation*}
can be evaluated via Fourier analysis to obtain an exact solution. We recognize that Equation \eqref{eq:gelu_moment_a} is a convolution in $\mu$ and exploit the Fourier convolution property. The Fourier transforms of interest are:
\begin{align*}
\frac{1}{2} \ \mu \ \text{erf}\left( \frac{\mu}{\sqrt{2} } \right) &\xrightarrow{\mathfrak{F}} \frac{- (4 \pi^2 \xi^2 + 1) \exp \{ -2 \pi^2 \xi^2 \}}{4 \pi^2 \xi^2} \\
\frac{1}{\sqrt{2\pi \sigma^2} } \exp\left\{ -\frac{\mu^{2}}{2\sigma^{2} } \right\}  &\xrightarrow{\mathfrak{F}} \exp\left\{ -2  \pi^{2} \xi^{2} \sigma^{2} \right\}.
\end{align*}

In the Fourier domain, our convolution becomes the product of the transforms:
\begin{equation*}
    - \frac{4 \pi^2 \xi^2 + 1}{4 \pi^2 \xi^2} \exp \{ -2 \pi^2 \xi^2 \} \exp\left\{ -2  \pi^{2} \xi^{2} \sigma^{2} \right\} \\
    = -\exp\left\{ -2  \pi^{2} \xi^{2} (1 + \sigma^{2}) \right\} - \frac{1}{4 \pi^2 \xi^2} \exp\left\{ -2  \pi^{2} \xi^{2} (1 + \sigma^{2}) \right\}.
\end{equation*}
This representation gives us two terms to which we can apply the inverse Fourier transform:
\begin{align*}
    -\exp\left\{ -2  \pi^{2} \xi^{2} (1 + \sigma^{2}) \right\} &\xrightarrow{\mathfrak{F}^{-1}} -\frac{1}{\sqrt{(1 + \sigma^2}} \phi\left(\frac{\mu}{\sqrt{1 + \sigma^{2}}} \right) \\
    -\frac{1}{4 \pi^2 \xi^2} \exp\left\{ -2  \pi^{2} \xi^{2} (1 + \sigma^{2}) \right\} &\xrightarrow{\mathfrak{F}^{-1}} \frac{1}{2} \mu \ \text{erf}\left( \frac{\mu}{\sqrt{2 (1 + \sigma^2)}} \right) + \sqrt{1 + \sigma^2} \phi \left(\frac{\mu}{\sqrt{1 + \sigma^{2}}} \right)
\end{align*}
Combining all results yields:
\begin{equation*}
\mathbb{E}[z] = \mu \Phi \left( \frac{\mu}{\sqrt{1 + \sigma^2}} \right) + \frac{\sigma^2}{\sqrt{1 + \sigma^2}} \phi \left( \frac{\mu}{\sqrt{1 + \sigma^2}} \right).
\end{equation*}
\end{proof}

Let $\alpha =\sigma /\sqrt{1+\sigma^{2} }$. We can rewrite Equation~\eqref{eq:gelu_mean_a} as
\begin{equation}\label{eq:gelu_mean2_a}
\mathbb{E}[z] =\mu\Phi \left( \frac{\alpha \mu}{\sigma} \right)  +\alpha \sigma \phi \left( \frac{\alpha \mu}{\sigma} \right).
\end{equation}

The variance of a GELU doesn't appear to have a closed-form solution, but can be expressed analytically with Corollary~\ref{thm:var}. To apply Theorem~\ref{thm:cov} or Corollary~\ref{thm:var}, the first few terms of interest are:
\begin{align*}
\sigma \frac{\partial \mathbb{E}[z] }{\partial \mu}    &=   \sigma \Phi \left( \frac{\alpha \mu}{\sigma} \right) + \alpha \left(1 - \alpha^2 \right) \mu  \phi \left( \frac{\alpha \mu}{ \sigma }\right) \\
\sigma^2 \frac{\partial^{2} \mathbb{E}[z] }{\partial \mu^{2}} &=  \alpha \left[  - \left( \frac{\alpha \mu}{\sigma} \right)^2  \left(1 - \alpha^2 \right) + \left(2 - \alpha^2 \right) \right] \sigma\  \phi \left( \frac{\alpha \mu}{ \sigma }\right) \\
\sigma^3 \frac{\partial^{3} \mathbb{E}[z] }{\partial \mu^{3}} &=  - \alpha^2\left[  - \left( \frac{ \alpha \mu}{\sigma} \right)^2  \left( 1 - \alpha^2 \right) + \left(4 - 3 \alpha^2 \right)\right]\sigma \left( \frac{\alpha \mu}{\sigma} \right)\phi \left( \frac{\alpha \mu}{ \sigma }\right) \\
\sigma^4 \frac{\partial^{4} \mathbb{E}[z] }{\partial \mu^{4}} &=  \sigma \alpha^3 \left[ - \left( \frac{\alpha \mu}{\sigma} \right)^4 \left( 1 - \alpha^2 \right) + \left( \frac{\alpha \mu}{\sigma} \right)^2 \left( 7 - 6 \alpha^2 \right) - \left( 4 - 3 \alpha^2 \right) \right] \phi \left( \frac{\alpha \mu}{ \sigma }\right) \\
\sigma^5 \frac{\partial^{5} \mathbb{E}[z] }{\partial \mu^{5}} &=  -\sigma \alpha^4  \left[ -\left(\frac{\alpha \mu}{\sigma} \right)^4 \left(1 - \alpha^2 \right) + \left(\frac{\alpha \mu}{\sigma} \right)^2 \left( 11 - 10 \alpha^2 \right) - 3 \left( 6 - 5 \alpha^2 \right) \right] \left(\frac{\alpha \mu}{\sigma}\right) \phi \left( \frac{\alpha \mu}{ \sigma }\right).
\end{align*}

With the probabilist's Hermite polynomial, we can express terms for $k>1$ as
\begin{equation}\label{eq:gelu_hermite_a}
\sigma^k \frac{\partial^{k} \mathbb{E}[z] }{\partial \mu^{n}} =  \alpha^{k-1} \sigma \  (-1)^k \ \left[  \mathrm{He_{k-2}} \left( \frac{\alpha \mu}{\sigma} \right) - \left( 1-\alpha^2 \right)  \mathrm{He_{k}} \left( \frac{\alpha \mu}{\sigma} \right)\right] \phi \left( \frac{\alpha \mu}{\sigma} \right).
\end{equation}
We note that as $\sigma \rightarrow \infty$, $\alpha \rightarrow 1$, Equation~\eqref{eq:gelu_mean2_a} reduces to Equation~\eqref{eq:mean_relu_a}, and Equation~\eqref{eq:gelu_hermite_a} reduces to Equation~\eqref{eq:relu_hermite_a}.

\subsection{Sigmoid}\label{sec:appendix_sigmoid}

Sigmoid functions frustrate many standard analytical tools, and moments of Gaussian random variables passed through sigmoid functions do not appear to have closed-form solutions. The logistic sigmoid is defined as
\begin{equation*}
    s(y) = \frac{1}{1 + e^{-y}}.
\end{equation*}
For a Gaussian input $y \sim \mathcal{N}(\mu, \sigma^2)$, \cite{daunizeau_semi-analytical_2017} proposes the following approximation for the mean of $z = s(y)$:
\begin{equation}\label{eq:sigmoid_daunizeau}
    \mathbb{E}[z] \approx s\left( \frac{\mu}{\sqrt{1 + \alpha \sigma^2}} \right)
\end{equation}
where $\alpha = 0.368$ is an empirically derived parameter. We note in passing that the best-fit $\alpha$ is sensitive to the range of considered input means and variances. For example, we find that $\alpha = \pi/8$, derived from $\tanh(y) \approx \text{erf}\left( \frac{\sqrt{\pi}}{2} y \right)$, is sometimes a better fit.

Let $\beta = \sqrt{1 + \alpha \sigma^2}$. To compute the approximate covariance of the output of a sigmoid function using Theorem~\ref{thm:cov} and Equation~\eqref{eq:sigmoid_daunizeau}, the first few terms of interest are:
\begin{align*}
\sigma \frac{\partial \hat{\mathbb{E}}[z] }{\partial \mu} &= \sigma \frac{e^{\mu/\beta}}{\beta \left(1 + e^{\mu/\beta} \right)^2} \\
\sigma^2 \frac{\partial^{2} \hat{\mathbb{E}}[z] }{\partial \mu^{2}} &= \sigma^2 \frac{e^{\mu/\beta} \left( -1 + e^{\mu/\beta} \right)}{\beta^2 \left(1 + e^{\mu/\beta} \right)^3} \\
\sigma^3 \frac{\partial^{3} \hat{\mathbb{E}}[z] }{\partial \mu^{3}} &= \sigma^3 \frac{e^{\mu/\beta} \left( 1 - 4e^{\mu/\beta} + e^{2\mu/\beta} \right)}{\beta^3 \left(1 + e^{\mu/\beta} \right)^4} \\
\sigma^4 \frac{\partial^{4} \hat{\mathbb{E}}[z] }{\partial \mu^{4}} &= \sigma^4 \frac{e^{\mu/\beta} \left( -1 + 11e^{\mu/\beta} - 11e^{2\mu/\beta} + e^{3\mu/\beta} \right)}{\beta^4 \left(1 + e^{\mu/\beta} \right)^5} \\
\sigma^5 \frac{\partial^{5} \hat{\mathbb{E}}[z] }{\partial \mu^{5}} &= \sigma^5 \frac{e^{\mu/\beta} \left( 1 - 26e^{\mu/\beta} + 66e^{2\mu/\beta} - 26e^{3\mu/\beta} + e^{4\mu/\beta} \right)}{\beta^5 \left(1 + e^{\mu/\beta} \right)^6}.
\end{align*}

\section{ADDITIONAL EXPERIMENT DETAILS}

For the MNIST experiments discussed in Section~\ref{sec:trained_experiments}, the architecture of the trained CNN is summarized in Table~\ref{tab:mnist_layers}. As mentioned in Section~\ref{sec:trained_experiments}, we use strided, activation-less convolutional layers in place of pooling layers~\citep{springenberg2015striving} to apply moment propagation.

\begin{figure}[htbp]
     \centering
     \begin{subfigure}[b]{0.35\textwidth}
         \centering
         \includegraphics[width=\textwidth]{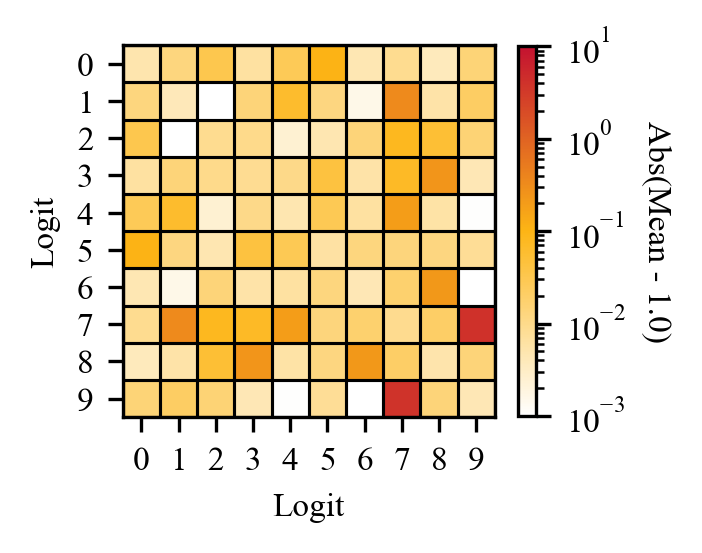}
         \caption{}
         \label{fig:cov_tightness_mean_mp}
     \end{subfigure}
     \begin{subfigure}[b]{0.35\textwidth}
         \centering
         \includegraphics[width=\textwidth]{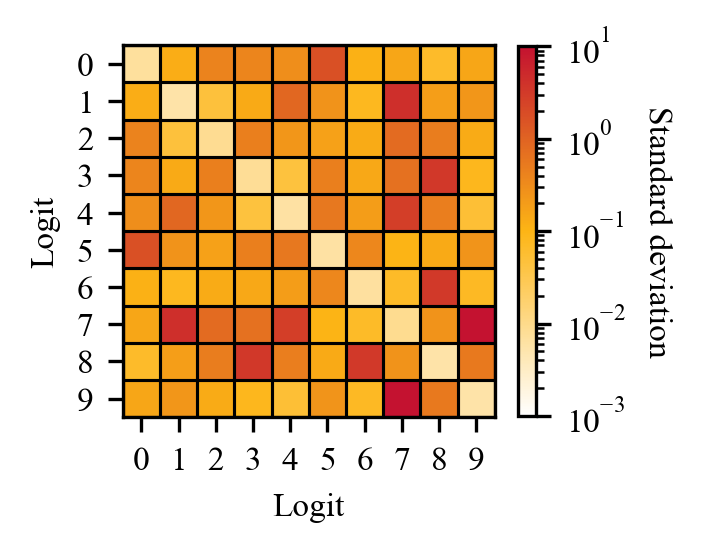}
         \caption{}
         \label{fig:cov_tightness_std_mp}
     \end{subfigure}
     \begin{subfigure}[b]{0.35\textwidth}
         \centering
         \includegraphics[width=\textwidth]{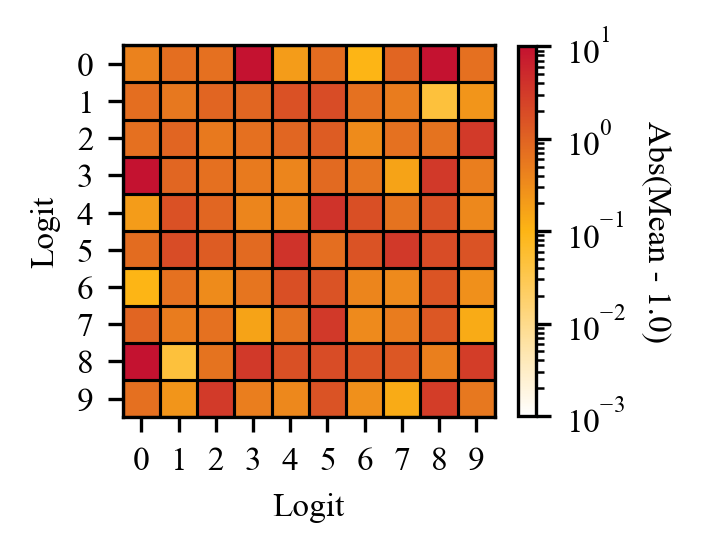}
         \caption{}
         \label{fig:cov_tightness_mean_pl}
     \end{subfigure}
     \begin{subfigure}[b]{0.35\textwidth}
         \centering
         \includegraphics[width=\textwidth]{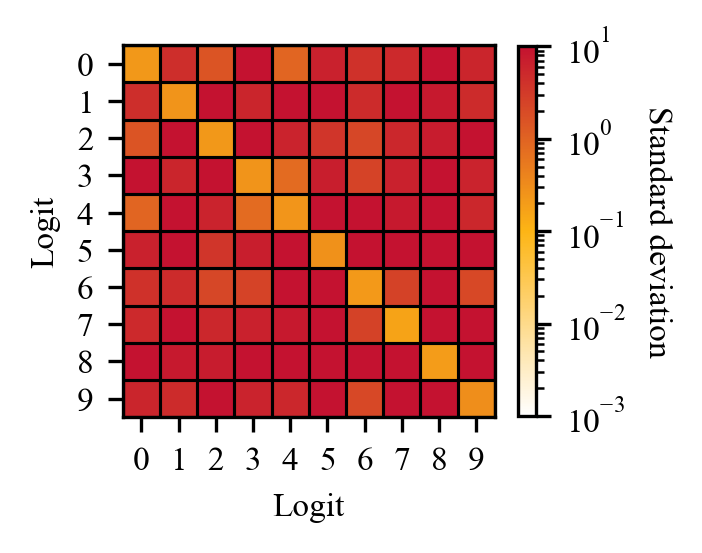}
         \caption{}
         \label{fig:cov_tightness_std_pl}
     \end{subfigure}
     \caption{Tightness measures of covariance for a convolutional network trained on MNIST data, comparing the proposed moment propagation method to PL-DNN~\citep{bibi2018}. Complementary to the ratio of variance estimates $\Qoppa{\hat{\sigma}^2}=\hat{\sigma}^2_{MC}/\hat{\sigma}^2_{A}$ in Table~\ref{tab:tightness_mnist}, for the 10 class logits we visualize the (a) mean and (b) standard deviation of the ratio of covariance estimates using moment propagation, and the (c) mean and (d) standard deviation of the ratio of covariance estimates using PL-DNN, calculated over 200 instances. Lower is better.}
     \label{fig:cov_tightness_appendix}
\end{figure}

A visualization of the ratio of covariance estimates, complementing the ratio of variance estimates $\Qoppa{\hat{\sigma}^2}=\hat{\sigma}^2_{MC}/\hat{\sigma}^2_{A}$ in Table~\ref{tab:tightness_mnist}, is shown in Figure~\ref{fig:cov_tightness_appendix}. Overall, for both the proposed method and PL-DNN, the means are similar to the mean of the corresponding ratio of variance estimates, but the standard deviations are greater.

\begin{table}[htb]
    \caption{Model summary for MNIST tightness experiments.}
    \label{tab:mnist_layers}
    \label{tab:mnist_layers_strided}
    \centering
    \begin{tabular}{rllcl}
    \toprule
    Layer & Type & Stride & Output Size & Activation \\
    \midrule
    $1$ & Convolutional & - & $28 \times 28 \times 32$ & ReLU \\
    $2$ & Convolutional & $2$ & $14 \times 14 \times 32$ & None \\
    $3$ & Convolutional & - & $14\times 14 \times 64$ & ReLU \\
    $4$ & Convolutional & $2$ & $7 \times 7 \times 64$ & None \\
    $5$ & Fully Connected & - & $1024$ & ReLU \\
    $6$ & Fully Connected & - & $10$ & Softmax \\
    \bottomrule
    \end{tabular}
\end{table}

For the UCI regression experiments discussed in Section~\ref{sec:bnn_experiments}, the considered hyperparameters are shown in Table~\ref{tab:bnn_hyperparameters}. As noted in Section~\ref{sec:bnn_experiments}, the hidden-layer width and number of training epochs is $50$ except for the larger \texttt{protein} dataset, for which we use $100$ hidden units and and train over $25$ epochs.

\begin{table}[htbp]
    \centering
    \caption{BNN hyperparameters considered for UCI regression tasks.}
    \label{tab:bnn_hyperparameters}
    \begin{tabular}{ll}
    \toprule
    Hyperparameter & Value \\
    \midrule
    Activation & ReLU \\
    Batch size & $10$ \\
    Hidden layers & 1 \\
    Hidden-layer width & $50$* \\
    Lambda & $1\mathrm{e}{-3}$ \\
    Learning rate & $\{1\mathrm{e}{-3}, 3\mathrm{e}{-3}, 5\mathrm{e}{-3}, 7\mathrm{e}{-3}, 1\mathrm{e}{-2}\}$ \\
    Optimizer & AdamW \\
    Training epochs & $50$* \\
    \bottomrule
    \end{tabular}
\end{table}

\end{document}